\documentclass[accepted]{uai2021} 

\usepackage[american]{babel}

\usepackage{natbib} 
\usepackage{mathtools} 
\usepackage{booktabs} 
\usepackage{tikz} 

\usepackage{times}
\usepackage{epsfig}
\usepackage{graphicx}
\usepackage{amsmath}
\usepackage{amssymb}
\usepackage{amsthm}
\usepackage{physics}
\usepackage{xr}

\makeatletter
\newcommand*{\addFileDependency}[1]{
  \typeout{(#1)}
  \@addtofilelist{#1}
  \IfFileExists{#1}{}{\typeout{No file #1.}}
}
\makeatother

\newtheorem{theorem}{Theorem}[section]

\theoremstyle{remark}

\theoremstyle{definition}
\newtheorem{definition}{Definition}[section]

\usepackage{algorithm}
\usepackage{algorithmic}
\usepackage[caption=false]{subfig}

\title{Staying in Shape: Learning Invariant Shape Representations using Contrastive Learning}

\author[1]{\href{mailto:Jeffrey Gu <jeffgy@stanford.edu>?Subject=Your UAI 2021 paper}{Jeffrey~Gu}{}}
\author[2]{Serena Yeung}
\affil[1]{%
    Institute for Computational \& Mathematical Eng.\\
    Stanford University\\
    Stanford, California, USA
}
\affil[2]{%
    Depts. of Biomedical Data Science and Computer Science\\
    Stanford University\\
    Stanford, California, USA
}

\begin{document}
\maketitle

\begin{abstract}
  Creating representations of shapes that are invariant to isometric or almost-isometric transformations has long been an area of interest in shape analysis, since enforcing invariance allows the learning of more effective and robust shape representations. Most existing invariant shape representations are handcrafted, and previous work on learning shape representations do not focus on producing invariant representations. 
  To solve the problem of learning unsupervised invariant shape representations, we use contrastive learning, which produces discriminative representations through learning invariance to user-specified data augmentations. To produce representations that are specifically isometry and almost-isometry invariant, we propose new data augmentations that randomly sample these transformations. We show experimentally that our method outperforms previous unsupervised learning approaches in both effectiveness and robustness. 
\end{abstract}

\section{Introduction}

3D shape analysis is important for many applications, such as processing street-view data for autonomous driving \citep{pylvanainen2010automatic}, studying morphological differences arising from disease \citep{niethammer2007global}, archaeology \citep{richards2012kinect}, and virtual reality \citep{hagbi2010shape}. Deep learning methods for shape analysis have generally focused on the supervised setting. However, manual annotations are expensive and time-consuming to produce in 3D. In some cases, annotations may even be impossible to produce, for example in biomedical imaging, where annotating pathological specimens may be hindered by a limited understanding of the disease. Unsupervised learning allows us to avoid the need to produce manual annotations. 

3D data comes in many formats, each of which has advantages and disadvantages, and their own methods for shape analysis. Voxel data consists of a 3D grid of voxels, but tends to suffer from data sparsity, low voxel resolution, and shape learning methods tend to be computationally expensive \citep{wei2020view}.  Point cloud data consists of a list of coordinates representing points on the shape, and is generally more dense than voxel data and also more easily permits direct transformations on the shape represented by the data. Because of these reasons, we will focus on point cloud data in our paper.   

Previous unsupervised methods for learning shape descriptors have generally used either probabilistic models \citep{xie2018learning, shi2020unsupervised}, generative adversarial networks (GANs) \citep{wu20153d, achlioptas2018learning, han2019view}, or autoencoders \citep{girdhar2016learning, sharma2016vconv, wu20153d, yang2018foldingnet}. One approach that has been relatively unexplored for deep learning methods but common in hand-crafted methods is to design shape descriptors that are invariant to transforms that preserve distances, either the extrinsic (Euclidean) distance \citep{belongie2001shape, johnson1999using, manay2004integral, gelfand2005robust, pauly2003multi} or intrinsic (geodesic) distance \citep{elad2003bending, rustamov2007laplace, sun2009concise, aubry2011wave}. Distance-preserving transformations are called isometries, and such transformations preserve only the underlying shape properties. In this paper, we will focus on extrinsic isometries, which include many common transformations such as rotations, reflections, and translations. Enforcing isometry-invariance leads to more effective representations by simplifying the learning problem for downstream tasks, since we will only need to learn the task for each possible shape and not each possible example. Furthermore, invariance also makes our learned representations robust to the variation in shapes. However, isometry-invariance is unable to distinguish between different poses of a shape, such as a when an object bends. These poses are instead almost-isometric, and we argue that almost-isometry invariance can capture these cases while retaining the benefits of isometry-invariance. 


To learn isometry and almost-isometry invariant representations, we use contrastive learning in combination with methods that sample isometric and almost-isometric transformations to learn invariant representations in an unsupervised fashion. Contrastive learning allows the learning of representations that are both invariant and discriminative \citep{xiao2020should} through the use of instance discrimination as a pretext task, where the model is trained to match an input to its transformed or augmented version. However, existing isometric data augmentation methods such as random rotation around the gravity axis, which were originally proposed for supervised point cloud learning, are not general enough to achieve our goal of learning invariance to general extrinsic isometries or almost-isometries. To do this, we introduce novel data augmentations that are capable of sampling general isometries and almost-isometries using mathematical results on sampling from groups, for isometries, and concentration of measure, for linear almost-isometries. We also propose a new smooth perturbation augmentation to capture additional non-linear isometries. 

Our focus on learning transformation-invariant representations also leads to more robust representations. Robustness is useful for real-world applications where the data may be noisy or have arbitrary orientation or pose, and may also offer greater protection against adversarial attacks \citep{zhao2020isometry}. However, few previous unsupervised shape representation learning methods have investigated the robustness of their methods, and those that do observe drop-offs in performance on downstream tasks as the noise level increases. Our invariance-based method is able to overcome these limitations. 

We show empirically that previous point cloud data augmentations are insufficient for learning good representations with contrastive learning, whereas our proposed data augmentations result in much more effective representations. We also show the quality of representations learned with contrastive learning and our new data augmentations for downstream shape classification. Finally, we demonstrate that our representations are also more robust to variations such as rotations and perturbations than previous unsupervised work. 


\section{Related Works}

\paragraph{Shape Descriptors}
Shape descriptors represent 3D shapes as a compact $d$-dimensional vector with the goal of capturing the underlying geometric information of the shape. Many hand-crafted shape descriptors have focused on enforcing invariance to various types of isometries, such as extrinsic isometries (i.e. isometries in Euclidean space) \citep{belongie2001shape, johnson1999using, manay2004integral, gelfand2005robust, pauly2003multi} or isometries intrinsic to the shape itself \citep{rustamov2007laplace, sun2009concise, aubry2011wave}.

Unsupervised methods for learning shape descriptors follow two major lines of research, with the first line leveraging generative models such as autoencoders \citep{girdhar2016learning, sharma2016vconv, yang2018foldingnet} or generative adversarial networks (GANs) \cite{wu2016learning, achlioptas2018learning, han2019view} and the second line focusing on probabilistic models \citep{xie2018learning, shi2020unsupervised}. Autoencoder-based approaches focus either on adding additional supervision to the latent space via 2D predictability \citep{girdhar2016learning}, adding de-noising \citep{sharma2016vconv}, or improving the decoder using a folding-inspired architecture \citep{yang2018foldingnet}. GAN-based approaches leverage either an additional VAE structure \citep{wu2016learning}, pre-training via earthmover or Chamfer distance \citep{achlioptas2018learning}, or using inter-view prediction as a pretext task \citep{han2019view}. For probabilistic methods, \citet{xie2018learning} proposes an energy-based convolutional network which is trained with Markov Chain Monte Carlo such as Langevin dynamics, and \citet{shi2020unsupervised} proposes to model point clouds using a Gaussian distribution for each point. Of these approaches, only \citet{shi2020unsupervised} focuses on producing robust representations.

Finally, some methods do not fall under any of these three approaches. \citet{sauder2019self} uses reconstruction as a pretext task to self-supervise representation learning. 
PointContrast \citep{xie2020pointcontrast} aims to learn per-point representations using a novel residual U-Net point cloud encoder and a per-point version of InfoNCE \citep{oord2018representation}. They use contrastive learning to pre-train on views generated from ScanNet \citep{dai2017scannet}, a dataset of 3D indoor scenes. In contrast, our work focuses specifically on learning isometry and almost-isometry invariant representations of shapes and developing algorithms to sample such transformations. 

\paragraph{Contrastive Learning}

Contrastive learning has its roots in the idea of a pretext task, a popular approach in unsupervised or self-supervised learning. A pretext task is any task that is learned for the purpose of producing a good representation \citep{he2020momentum}. Examples of pretext tasks for 2D image and video data include finding the relative position of two patches sampled from an image \citep{doersch2015unsupervised}, colorizing grayscale images \citep{zhang2016colorful}, solving jigsaw puzzles \citep{noroozi2016unsupervised}, filling in missing patches of an image \citep{pathak2016context}, and predicting which pixels in a frame of a video will move in subsequent frames \citep{pathak2017learning}. Contrastive learning can be thought of as a pretext task where the goal is to maximize representation similarity of an input query between positive keys and dissimilarity between negative keys. Positive keys are generated with a stochastic data augmentation module which, given an input, produces a pair of random views of the input \citep{xiao2020should}. The other inputs in the batch usually serve as the negative keys. The main application of contrastive learning has been to learn unsupervised representations of 2D natural images \citep{chen2020simple, he2020momentum, chen2020improved, xiao2020should}. We focus on using contrastive learning as an means of producing shape-specific invariant representations for 3D point clouds. 


\paragraph{Data Augmentation}

Although data augmentation has been well-studied for 2D image data, there has been little work studying data augmentations for point clouds. Previously examined point cloud augmentations include rotations around the the gravity axis, random jittering, random scaling, and translation \citep{qi2017pointnet, qi2017pointnet++, li2020pointaugment} in the supervised learning setting, and applying a random rotation from 0 to 360$^\circ$ on a randomly chosen axis for unsupervised pre-training \citep{xie2020pointcontrast}. \citet{chen2020pointmixup} proposes to generalize image interpolation data augmentation to point clouds using shortest-path interpolation. To improve upon these hand-crafted data augmentations, \citet{li2020pointaugment} proposes an auto-augmentation framework that jointly optimizes the data augmentations and a classification neural network, but is not applicable in unsupervised settings. In contrast, our work focuses on generalizing previous data augmentations such as random rotation and jittering to much more general classes of invariant transformations, including Euclidean isometries and almost-isometries, for the purpose of invariant representation learning with contrastive learning. 

\section{Methods}

In this section, we introduce our novel transformation sampling schemes and the contrastive learning framework we use to learn invariant representations. In Section \ref{augmentations}, we introduce sampling procedures for isometry and almost-isometry invariant transformations, and in Section \ref{contrastivelearning} we show how contrastive learning can be used to learn representations that are invariant to the transformations introduced in Section \ref{augmentations}.  

\subsection{Sampling isometric and almost-isometric transformations}\label{augmentations}

\begin{figure*}[ht]
\vskip 0.2in
\begin{center}
\begin{minipage}{\linewidth}
    \subfloat[]{\includegraphics[width=0.25\linewidth] {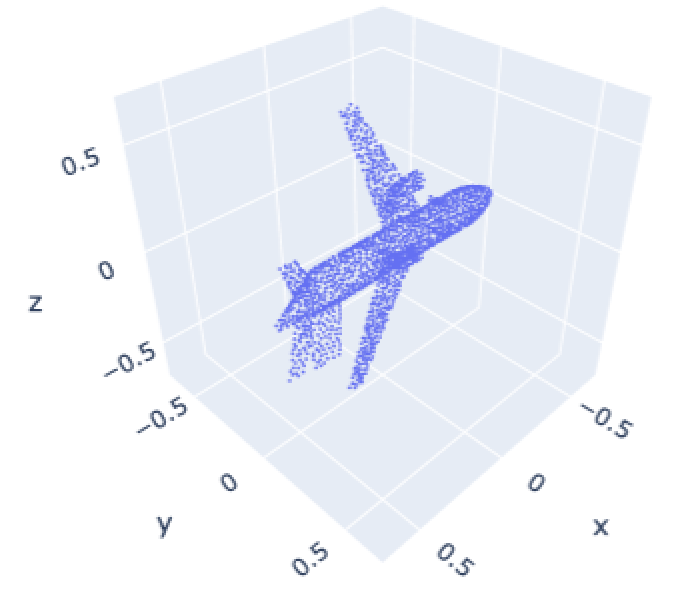}}
    \subfloat[]{\includegraphics[width=0.25\linewidth] {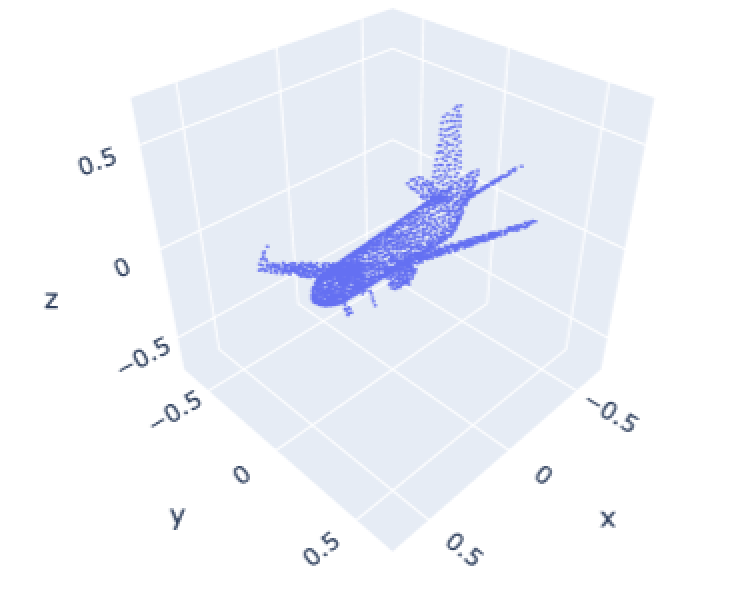}}
    \subfloat[]{\includegraphics[width=0.25 \linewidth] {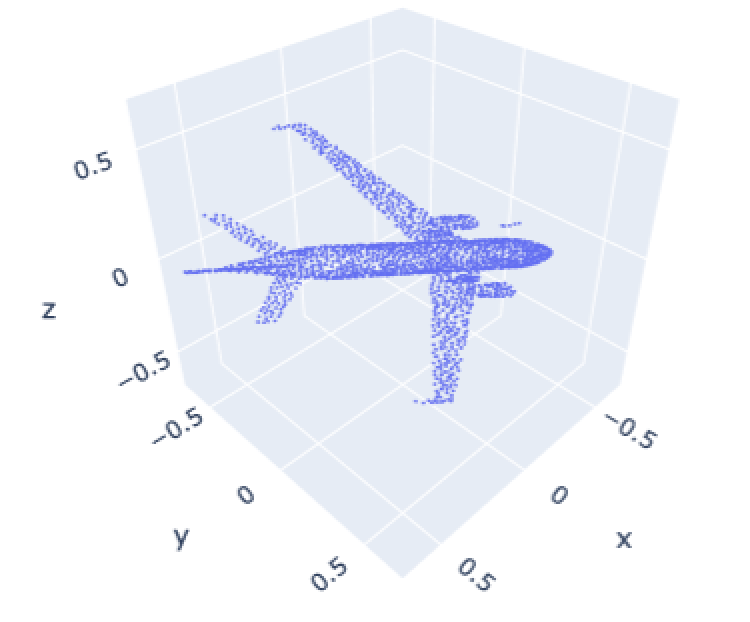}} 
    \subfloat[]{\includegraphics[width=0.25\linewidth] {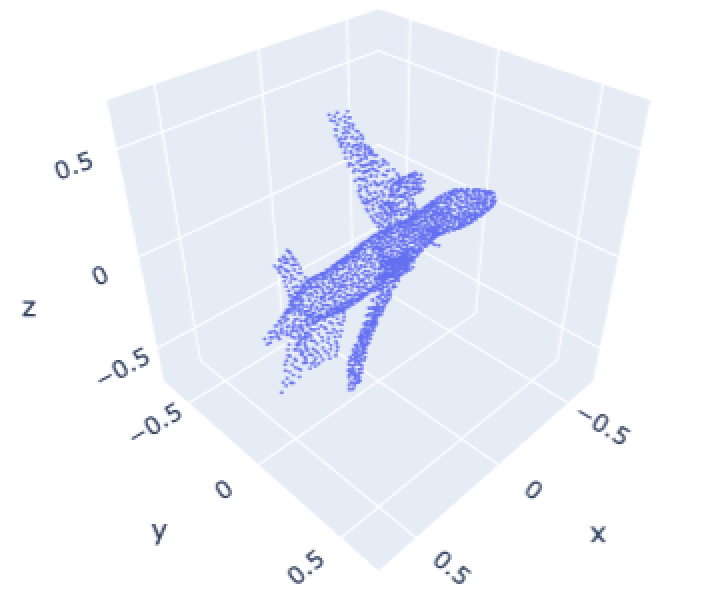}} 
\end{minipage}
\caption{Examples of our isometric and almost-isometric transformations. Each image has been normalized to be centered at the origin and scaled so the maximum distance of any point to the origin is 1. (a): The original point cloud. (b): The point cloud after a uniformly sampled orthogonal transform has been applied. We see that the point cloud has been rotated. (c): The point cloud after a random RIP transformation has been applied. The point cloud has undergone both rotation and a small amount of stretching (d): The point cloud after a smooth perturbation has been applied. We see that the point cloud has been perturbed, particularly near the nose of the aircraft.}
\label{fig:examples}
\end{center}
\vskip -0.2in
\end{figure*}

To achieve our goal of learning isometry-invariant and almost-isometry-invariant representations, we develop algorithms that allow us to sample randomly instances of these transformations from the set of all such transformations.

\paragraph{Preliminaries}

An isometry is a distance-distance preserving transformation:
\begin{definition}
Let $X$ and $Y$ be metric spaces with metrics $d_X, d_Y$. A map $f: X \to Y$ is called an isometry if for any $a, b \in X$ we have $d_X(a, b) = d_Y(f(a), f(b))$.
\end{definition}
In this paper, we will only be concerned about isometries of Euclidean space ($X = Y = \mathbb{R}^n$). Examples of Euclidean isometries include translations, rotations, and reflections. Mathematically, if two objects are isometric, then the two objects are the same shape.
From a shape learning perspective, isometry-invariance creates better representations by allowing downstream tasks such as classification to learn only one label per shape, rather than having to learn the label of every training example. 

\subsubsection{Uniform orthogonal transformation}
\label{orth}

The isometries of $n$-dimensional Euclidean space are described by the Euclidean group $E(n)$, the elements of which are arbitrary combinations of rotations, reflections, and translations. If we normalize each point cloud by centering it at the origin, then we only need to consider linear isometries, which are precisely the orthogonal matrices $O(n)$ (for more details, see Appendix~\ref{orth-info}). In the rest of the paper, we will use orthogonal transformation and isometry interchangeably. 

To ensure robustness to all orthogonal transformations $Q \in O(n)$, we would like to sample uniformly $Q$ from $O(n)$. A biased sampling method may leave our algorithm with ``blind spots'', as it may only learn to be invariant to the more commonly sampled orthogonal transformations. 
A theorem of Eaton \citep{eaton1983} shows that if a random matrix $A$ whose entries are distributed according to the standard normal distribution is QR-factorized, then $Q$ distributed uniformly on $O(n)$. This provides a simple algorithm for sampling uniform orthogonal transformations, given in Algorithm \ref{alg:uniform-orth}. An example transformation is shown in Figure \ref{fig:examples}. 

\begin{algorithm} 
\caption{Uniform Orthogonal sampling}\label{alg:uniform-orth}
\begin{algorithmic}[1]
\REQUIRE dimension $n$
\ENSURE uniform orthogonal matrix $Q \in O(n)$
\STATE Sample $A \sim N(0, 1)^{n \times k}$
\STATE Perform QR decomposition on $A$ to get $Q, R$
\STATE \textbf{return} $Q$
\end{algorithmic}
\end{algorithm}

\subsubsection{Random almost-orthogonal transformation}
\label{rip}
Many transformations preserve almost all shape information but may not be isometries. For example, the bending of a shape or rotation of part of a shape around a joint generally change geodesic distances on the shape very little and are thus almost-isometric transformations. Using almost-isometries instead of exact isometries may also allow our shape representations to account for natural variation or small amounts of noise between two shapes that otherwise belong to the same class of shape. 

In the case of Euclidean isometries, an almost-isometric transformation is an almost-orthogonal transformation. To formally define almost-orthgonal matrices, we use the Restricted Isometry Property (RIP) first introduced by \citet{candes2005decoding}: 


\begin{definition}[Restricted Isometry Property of \citet{baraniuk2008simple}]
A $n \times N$ matrix $A$ satisfies the \textit{Restricted Isometry Property} of order $k$ if there exists a $\delta_k \in (0, 1)$ such that for all sets of column indices $T$ satisfying that $|T| \le k$ we have
\begin{align}
    (1 - \delta_k) \norm{x_T}^2 \le \norm{A_T x_T}^2 \le (1 + \delta_k)\norm{x_T}^2
\end{align}
where $A_T$ is the $n \times |T|$ matrix generated by taking columns of $A$ indexed by $T$, and $x_T$ is the vector obtained by retaining only the entries corresponding to the column indices $T$, and $N$ is an arbitrary parameter satisfying $N \gg n$.   
\end{definition}

For more details on RIP matrices, see Appendix \ref{rip-info}. To sample from the set of RIP matrices, we leverage the concentration of measure result of \citet{baraniuk2008simple} to create rejection sampling algorithm: 

\begin{theorem}[Theorem 5.2 of \citet{baraniuk2008simple}]
Suppose that $n, N$ and $0 < \delta < 1$ are given. If the probability distribution generating the $n \times N$ matrices $A$ satisfies the concentration inequality
\begin{align}
    \Pr(\left| \norm{Ax}^2 - \norm{x}^2 \right| \ge \epsilon \norm{x}^2) \le 2\epsilon^{-n c_0(\epsilon)}
\end{align}
where $0 < \epsilon < 1$ and $c_0$ is a constant depending only on $\epsilon$, then there exist constants $c_1, c_2 > 0$ depending only on $\delta$ such that RIP holds for $A$ with the prescribed $\delta$ and any $k \le c_1 n / \log(N/k)$ with probability $\ge 1 - e^{-c_2 n}$. 
\end{theorem}
We note that many common distributions satisfy the concentration inequality, for example $A_{ij} \sim \mathcal{N} \left(0, \frac{1}{n} \right)$ \cite{baraniuk2008simple}, where the concentration inequality holds with $c_0(\epsilon) = \epsilon^2/4 - \epsilon^3/6$. 

This theorem says that with the right setting of parameters, if we generate a random $n \times N$ matrix $A$ where the entries are chosen from a distribution satisfying the concentration inequality and form a new matrix $Q$ by taking $T$ random columns of $A$, the result is an $n \times T$ RIP matrix with high probability. This gives us a simple algorithm for sampling RIP matrices: first we generate a random matrix $A$ by sampling entries from $\mathcal{N} \left(0, \frac{1}{n} \right)$, choosing $T$ columns of $A$ without replacement and forming a new matrix $Q$ consisting of just these columns, and testing if the matrix is RIP (that is, it satisfies Equation \ref{eqn:rip-spectral-cond}, see Appendix \ref{rip-info}), repeating the procedure if $Q$ is not RIP. The full algorithm is given in Algorithm \ref{alg:rand-rip}, and an example RIP transformation is shown in Figure \ref{fig:examples}. 

\begin{algorithm}
\caption{Sample $Q$ such that $\sigma(Q^TQ - I) < \delta$} \label{alg:rand-rip}
\begin{algorithmic}[1]
\REQUIRE dimensions $n, N, T$, tolerance $\delta$
\ENSURE $n \times T$ matrix $Q$ satisfying RIP
\WHILE{$\norm{Q^TQ - I_n}_2 > \delta$} 
\STATE Sample $A \sim \mathcal{N}\left(0, \frac{1}{n}\right)^{n \times N}$
\STATE Randomly choose $T$ columns of $A$ without replacement to get $n \times T$ matrix $Q$
\ENDWHILE
\STATE return $Q$
\end{algorithmic}
\end{algorithm}

\begin{figure*}[htb!]
\vskip 0.2in
\begin{center}
\includegraphics[width=0.8\linewidth]{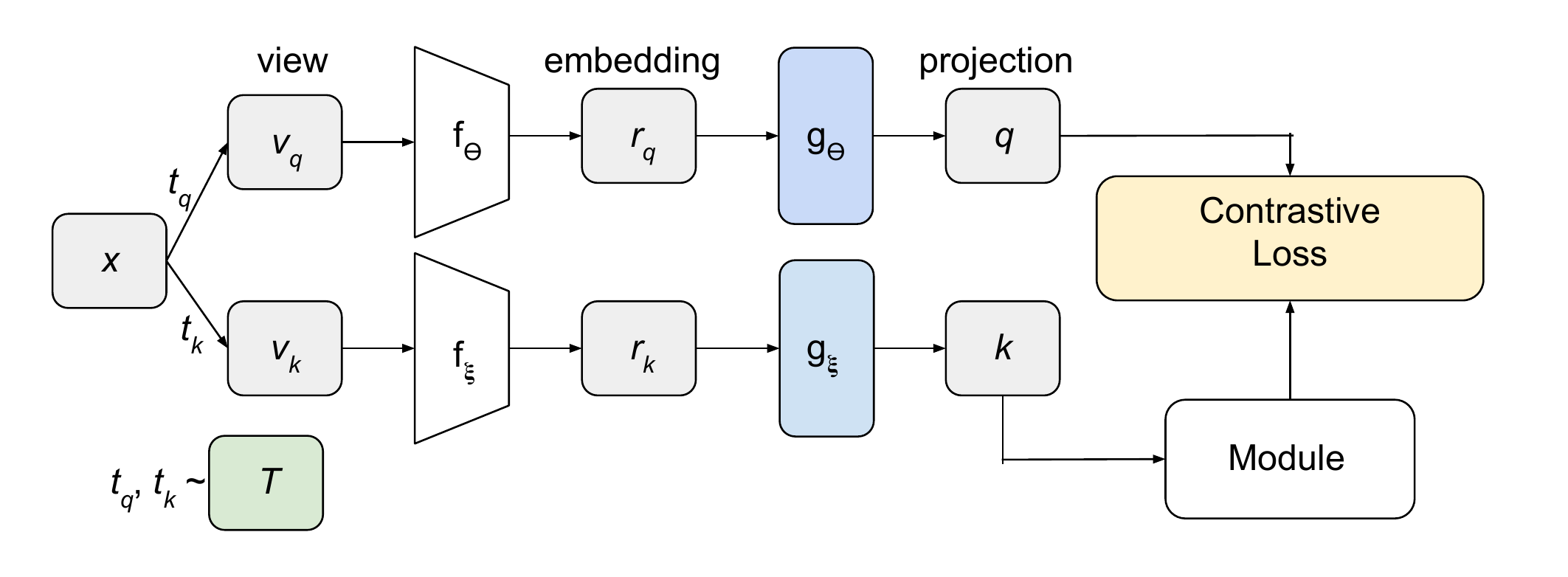}
\caption{Schematic of the contrastive learning framework as described in Section~\ref{contrastivelearning}. Random data augmentations $t_q, t_k$ are sampled from the stochastic data augmentation and applied to input $x$ to produce views $v_q, v_k$. The views are then fed through the corresponding encoder $f$ and then a projection head $g$ to produce representations $q, k$ which are then used to calculate the contrastive loss. The module block describes how the algorithm uses the key representations as negative examples. For example, in SimCLR \citep{chen2020simple}, the module is just the identity and the keys of all other views are used as negative examples, whereas MoCo \citep{he2020momentum, chen2020improved} uses a memory bank composed of key representations. Together, $,g(f(\cdot))$ comprise $E(\cdot)$. For methods employing a projection head $g$, for downstream tasks $g$ is thrown away and typically the representation $r_q$ is used.} 
\label{fig:contrastive-learning-diagram}
\end{center}
\vskip -0.2in
\end{figure*}

\subsubsection{Smooth perturbation}
\label{interp}

RIP transformations are examples of \textit{linear} almost-isometries, since they are represented by matrices. To capture some non-linear almost-isometries, we generalize the commonly used point cloud augmentation of Gaussian perturbation \citep{qi2017pointnet, qi2017pointnet++}, which applies Gaussian noise with zero mean to each point of the point cloud. To generalize this augmentation to capture the variation in real-world shapes, we propose a data augmentation that generates a smooth perturbation, inspired by \citep{ronneberger2015u, cciccek20163d}. We generate a smooth perturbation by sampling $P$ points uniformly in $\mathbb{R}^3$ and $3P$ values from a Gaussian with zero mean and standard deviation $\sigma$. We then use smooth interpolation to generate a perturbation $(n_x^i, n_y^i, n_z^i)$ for each point $p_i = (x_i, y_i, z_i)$ in the point cloud, and apply the perturbation as a translation of $p_i$ to get new points $p_i = (x_i + n_x^i, y_i + n_y^i, z_i + n_z^i)$. An example is shown in Figure \ref{fig:examples}. 

\subsection{Contrastive Learning}\label{contrastivelearning}

The contrastive learning framework (see Figure~\ref{fig:contrastive-learning-diagram}) can be summarized as follows \citep{xiao2020should}: we first define a stochastic data augmentation module $\mathcal{T}$ from which we can sample transformations $t \sim \mathcal{T}$. Given a training example $x$, two random views $v_q = t_q(x), v_k = t_k(x)$ are generated, where $t_q, t_k \sim \mathcal{T}$. We then produce representations $q, k$ by applying a base encoder $E(\cdot)$ to $v_q$ and $v_k$.  The pair $q, k_+ = k_1$ is called a positive pair, and our goal is to distinguish this pair from some set of negative examples $k_2, \ldots, k_K$. The model is then trained with a contrastive loss, which allows the model to learn representations that are invariant to the transformations in $\mathcal{T}$. 
We use InfoNCE \citep{oord2018representation} as our contrastive loss:
\begin{align}\label{infonce}
    \mathcal{L}_q = -\log \frac{\exp(q \cdot k_+/\tau)}{\sum_{i = 1}^K \exp(q \cdot k_i/\tau)}
\end{align}
where the temperature $\tau$ is a tunable hyperparameter. Since the contrastive loss forces $q, k_+$ to be similar and $q, k_i \neq k_+$ to be dissimilar, our model learns invariance to the transformations used to generate $q, k_+$. Many different strategies have been used to choose the negative keys $k_i \neq k_+$, such as using the keys of the other training examples in the mini batch \cite{chen2020simple} or drawing them from a queue of previously seen keys \cite{he2020momentum, chen2020simple}. 

We choose momentum contrastive learning (MoCo) \citep{he2020momentum, chen2020improved} as our contrastive learning framework due to its state-of-the-art performance for 2D image data and its relatively lightweight computational requirements, but our method is framework-agnostic and could be used with any contrastive learning framework. To adapt this framework for learning shape representations for point clouds, we need a base encoder capable of producing representations from point cloud input and shape-specific data transformations $T_i$. In our method, the stochastic data augmentation module $\mathcal{T}$ comprises the transformation-sampling modules introduced in Section \ref{augmentations}. Unlike the case of 2D image representations, where there are canonical choices of base encoder, there are not similar choices for point cloud data, due to the infancy of point cloud architectures \citep{xie2020pointcontrast}. PointNet \citep{qi2017pointnet}, DGCNN \citep{wang2019dynamic}, and a residual U-Net architecture \citep{xie2020pointcontrast} and others have all been used in prior work. Our framework is model-agnostic and works with any point cloud encoder. We will discuss the choice of base encoder more in Section \ref{experiments}. 

\section{Experiments} \label{experiments}

\subsection{Unsupervised Shape Classification Protocol}\label{shapeclassificationprotocol}

To show the quality of our learned shape representations, we compare our method to previous work on unsupervised shape classification. The procedure for our shape classification experiment follows the established protocol for unsupervised shape classification evaluation: first, the network is pre-trained in an unsupervised manner using the ShapeNet dataset \citep{chang2015shapenet}. Using the embeddings from pre-training, either a 2-layer MLP \citep{shi2020unsupervised} or linear SVM \citep{wu20153d} is trained and evaluated on the ModelNet40 dataset.
Following previous work \citep{wu20153d, shi2020unsupervised}, we only pre-train on the 7 major categories of ShapeNet (chairs, sofas, tables, boats, airplanes, rifles, and cars). Other work pre-train on all 55 categories of ShapeNet \citep{achlioptas2018learning, yang2018foldingnet, han2019view, sauder2019self}, but due to the differences in the amount of data used we are unable to make a fair comparison to these methods. 

\paragraph{ShapeNet}

ShapeNet \citep{chang2015shapenet} dataset consists of 57448 synthetic 3D CAD models organized into 55 categories with a further 203 subcategories, organized according to WordNet synsets. However, we only have access to the public version of ShapeNet, which contains the same categories but only 52472 models. For contrastive learning pre-training we use the normalized version of ShapeNet, where all shapes are consistently aligned and normalized to fit inside a unit cube. 

\paragraph{ModelNet40}

ModelNet40 \citep{wu20153d} is a shape classification dataset consisting of 12311 3D CAD models organized into 40 classes. We use the official ModelNet40 train and test splits of 9843 training examples of 2468 test examples. For downstream shape classification training and evaluation, we use the normalized and resampled version of ModelNet40, where models are normalized to be centered at the origin and and lie within the unit sphere and the points resampled as in \citet{qi2017pointnet}. ModelNet10 is a 10-class subset of ModelNet40. 

\paragraph{Training} We use PointNet \cite{qi2017pointnet} as our base encoder. For ShapeNet pre-training using MoCo, we follow \citet{he2020momentum, chen2020improved} and use SGD as our optimizer with weight decay 0.0001, momentum 0.9, temperature $\tau = 0.02$, and latent dimension 128. Unlike \citet{he2020momentum}, we train with only a single GPU with batch size 64 and a learning rate chosen from $\{0.075, 0.0075, 0.00075\}$, which is tuned using the final MoCo accuracy. Models are trained until the MoCo accuracy converges, up to a limit of 800 epochs. Convergence typically takes 200 epochs for single transformation models but up to or even exceeding 800 epochs for multiple transformation models. We use a cosine learning rate schedule \citep{chen2020simple, chen2020improved}. For both pre-training and supervised classification training, we sample 2048 points from each point cloud. 

For ModelNet40 shape classification we choose to use a two layer MLP, which is known to be equivalent to a linear SVM, and train with a batch size of 128, and a learning rate chosen from \{0.01, 0.001\}. The learning rate was selected using a validation set sampled from the official training set of ModelNet40. Following \citet{shi2020unsupervised}, our hidden layer has 1000 neurons.

\paragraph{Experimental setup} Unless otherwise stated, the setting of our data augmentation modules are as follows: for uniform orthogonal matrices, we set $n, k = 3$ to generate $3 \times 3$ orthogonal matrices. For random RIP matrices, we set $n = 3, N = 1000, T = 3$ and $\delta = 0.9$ (see Section \ref{rip}, Algorithm \ref{alg:rand-rip}). For the smooth perturbation data augmentation, we generate $100$ points according to an isotropic Gaussian with mean $0$ and standard deviation $0.02$, and perform radial basis interpolation to get smooth noise at every point in the point cloud, which we add to each point of the point cloud. For Gaussian noise, we perturb each point in the point cloud by a random perturbation sampled according to a Gaussian with mean $0$ and standard deviation $0.02$. 


\paragraph{Training with individual data augmentations}
Table \ref{table:single-aug} shows different versions of our method when trained with each individual transformation. We compare our proposed data augmentations against three existing data augmentations: random $y$-rotation \citep{qi2017pointnet}, random rotation \citep{zhao2020isometry}, and point cloud jitter/Gaussian perturbation \citep{qi2017pointnet}. We do not investigate random scaling or translations since their effect can always be negated by normalization. 

We first consider the linear transformations, which are the random $y$-rotation, random rotation from previous works and the uniform orthogonal transformation and random RIP transformations we propose. Each of the earlier classes of transformation is a subset of the later classes of transformations. We find that as the class of transformations get more general, the performance improves. This is similar to earlier contrastive learning work \citep{chen2020simple}, which finds that increasing the strength of a data augmentation improves the performance of contrastive learning. In particular, we find that the RIP transformation performs the best, followed by the uniform orthogonal transformation, showing that almost-isometry invariance provides further improvement over the more-strict isometry invariance.  We also find that our proposed transformations (uniform orthogonal, random RIP) greatly outperform previously used transformations for contrastive learning, and that these previous transformations are insufficient for learning good representations with contrastive learning (c.f. Table \ref{table:prior-work}). 

We find that the non-linear transformations (Gaussian perturbation and smooth perturbation) perform noticeably worse than the best linear transformations. We believe that this is because the best linear transformations captures more diversity in object variation. Both of the transformations in this category perform similarly, which is likely is due to the two transformations being similar in strength, since they are both based on noise sampled from a Gaussian distribution with the same standard deviation. 

\begin{table}[t]
\caption{Ablation study of our model pre-trained with only one transformation and on the 7 major ShapeNet categories listed in Section \ref{shapeclassificationprotocol} and evaluated using the protocol of Section \ref{shapeclassificationprotocol} on ModelNet40. Bolded names correspond to our proposed data augmentations.}
\vskip 0.15in
\begin{center}
\begin{small}
\begin{sc}
\begin{tabular}{c c c}
\hline
Type & Data augmentation & Accuracy \\
\hline
Linear & Random $y$-rotation & 71.8\% \\
& Random rotation & 72.9\%\\
& \textbf{Uniform Orthogonal} & 83.0\% \\
& \textbf{Random RIP} & 86.3\% \\
\hline
Non-linear & \textbf{Smooth perturbation} & 78.6\% \\
& Gaussian perturbation & 78.7\% \\
\hline
\end{tabular}
\end{sc}
\end{small}
\end{center}
\vskip -0.1in
\label{table:single-aug}
\end{table}

\paragraph{Training with multiple data augmentations}
\label{multiple-aug}

Previous contrastive learning literature finds that training with multiple transformations is generally more effective than training only a single transformation \citep{chen2020simple}, leading us to examine combinations of data augmentations. When training with multiple transformations, we uniformly randomly apply one of the transformations to each mini-batch. Due to the large number of combinations and the fact that many transformations are generalizations of other transformations, we only investigate the top two linear and non-linear transformations from Table \ref{table:single-aug}. Additionally, we only investigate all pairs of transformations. 

Table \ref{table:multiple-aug} shows the results of our method trained with pairs of data augmentation. Training was stopped for all models at 800 epochs regardless of whether the model was converged or not, due to the computational expense of training with single GPUs. Under these conditions, we find that the combination of the uniform orthogonal and random RIP transformations produces the best classification accuracy. We find that the random RIP and Gaussian perturbation and random RIP and smooth perturbation models do not fully converge after 800 epochs, in the sense that their instance discrimination accuracy after MoCo pre-training is still improving but not close to the accuracy achieved by the other models (above 90\%). 
In line with previous work, models trained with combinations of transformations improve over models trained with just the individual transformations in every case where the models converge. We conjecture that if computational resources were significantly increased, this would also hold for the models that have not converged, and for even greater combinations of data augmentations.  

\begin{table}[t]
\caption{Comparison of our model trained with combinations of augmentations mentioned in Section \ref{multiple-aug} and on the 7 major ShapeNet categories listed in Section \ref{shapeclassificationprotocol} and evaluated using the protocol of Section \ref{shapeclassificationprotocol} on ModelNet40. Here, orthogonal refers to our uniform orthogonal transformation, RIP refers to our random RIP transformation, perturbation refers to Gaussian perturbation, interpolation refers to our smooth perturbation generated using interpolation. 
Bolded names correspond to our proposed data augmentations. Models that did not converge after training with terminated at the maximum number of epochs (800) are marked with a $*$.}
\vskip 0.15in
\begin{center}
\begin{small}
\begin{sc}
\begin{tabular}{c c}
\hline
Data augmentations & Accuracy \\
\hline
\textbf{RIP} + \textbf{Interpolation}$^*$ & 73.0\% \\
\textbf{RIP} + Perturbation$^*$ & 75.9\% \\
\textbf{Orthogonal} + \textbf{Interpolation} & 83.6\% \\
\textbf{Orthogonal} + Perturbation & 83.9\% \\
Perturbation + \textbf{Interpolation} & 84.4\%\\
\textbf{Orthogonal} + \textbf{RIP} & 86.4\%\\
\hline
\end{tabular}
\end{sc}
\end{small}
\end{center}
\vskip -0.1in
\label{table:multiple-aug}
\end{table}

\paragraph{Comparison to previous results} Table \ref{table:prior-work} shows the performance of our method compared to previous unsupervised shape classification methods using the shape classification protocol. In the table, ``Ours'' refers to our model trained with the uniform orthogonal and random RIP transformations. 

Our model outperforms all comparable prior unsupervised work.
This shows the importance of learning invariance to shape-preserving transformations in shape representation learning, as no previous unsupervised methods explicitly consider learning invariant representations, as well as the importance of considering broadly invariant transformations in contrastive learning. Since most of the classes are unseen by the model during ShapeNet pre-training, our model also shows good ability to generalize to novel classes. 




\begin{table*}[ht]
\caption{Comparison of our method against previous unsupervised work on the shape classification protocol of Section \ref{shapeclassificationprotocol}. The evaluation metric is classification accuracy, and MN40 and MN10 refer to the ModelNet40 and ModelNet10 datasets, respectively. A $-$ indicates that there is no published result for that dataset.}
\vskip 0.15in
\begin{center}
\begin{small}
\begin{sc}
\begin{tabular}{c c c c}
\hline
Supervision & Method & MN40 & MN10\\
\hline
Supervised & PointNet \citep{qi2017pointnet} & 89.2\% &-- \\
 & PointNet++ \citep{qi2017pointnet++} & 91.9\% &--\\
 & PointCNN \citep{li2018pointcnn} & 92.2\% &--\\
 & DGCNN \citep{wang2019dynamic} & 92.2\% &--\\
 & RS-CNN \citep{liu2019relation} & 93.6\% &--\\
\hline
Unsupervised & T-L Network \citep{girdhar2016learning} & 74.4\% &-- \\
 & VConv-DAE \citet{sharma2016vconv} & 75.5\% & 81.5\%\\
 & 3D-GAN \citep{wu2016learning} & 83.3\% & 91.0\%\\
 & Point Distribution Learning \citep{shi2020unsupervised} & 84.7\% & -- \\
 & \textbf{Ours} & \textbf{86.4}\% & \textbf{92.8}\%\\
\hline
\end{tabular}
\end{sc}
\end{small}
\end{center}
\vskip -0.1in
\label{table:prior-work}
\end{table*}

\subsection{Robustness}
\label{robustness}


Our focus on learning transformation-invariant representations also leads to better representation robustness. Robust representations allow our method to better handle the natural variation in shapes and is useful in real-world settings where the input shapes may not always be consistently aligned. Additionally, robustness may also make our method more resistant to adversarial attacks. In this section, we assess robustness to common changes such as rotation and noise as well as more complex transformations based on our proposed data augmentations.

\begin{figure*}[htb!]
\vskip 0.2in
\begin{center}
\begin{minipage}{\linewidth}
    \subfloat[]{\includegraphics[width=0.25 \linewidth]{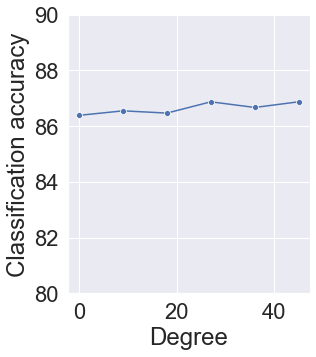}}
    \subfloat[]{\includegraphics[width=0.25 \linewidth]{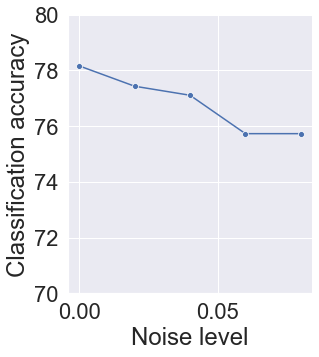}}
    \subfloat[]{\includegraphics[width=0.25 \linewidth]{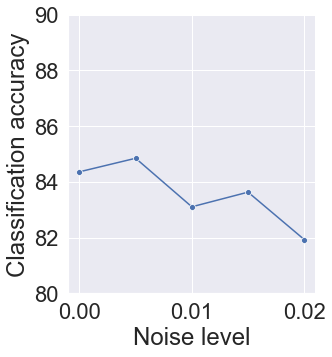}}
    \subfloat[]{\includegraphics[width=0.25 \linewidth]{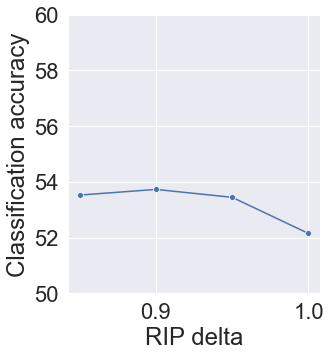}}
\end{minipage}
\caption{Plots of accuracy vs variation strength for (a) rotations by a fixed angle, (b) Gaussian noise of varying standard deviations, (c) smooth noise generated using Gaussian noise of varying standard deviations, and (d) RIP transformations with increasing deviation $\delta$ from isometry. Each variation was applied at both train and test time for ModelNet40 shape classification (see Section \ref{shapeclassificationprotocol}). We find that our method is fairly consistent with regards to different types of variation, with performance only decreasing slightly as the variation or noise becomes stronger.}
\label{fig:robustness}
\end{center}
\vskip -0.2in
\end{figure*}

\paragraph{Experimental Setup} 
In our first experiment, we examine robustness to rotation. Robustness to rotation can alleviate the need to align shapes before performing downstream tasks as well as provide greater defense against adversarial attacks \citep{zhao2020isometry}. We apply a rotation along each axis from 0 to 45 degrees in increments of 9 degrees to each shape during both supervised classification training and testing, following \citet{shi2020unsupervised}. All other experiment details are the same as Section \ref{shapeclassificationprotocol}. For this experiment, our model is trained with the uniform orthogonal and random RIP transformations.

As a second experiment, we evaluate the resistance of our method to noise, which is useful in real-world settings due to the imprecision of sensors. For this experiment, we apply a Gaussian perturbation with standard deviation 0 to 0.08 in increments of 0.02, and train our model with only the Gaussian perturbation with standard deviation 0.08. 

Finally, we evaluate robustness with respect to more complex variations such as the data augmentations proposed in this work. We show that our model is also robust to our proposed transformations, which are much more difficult than fixed-degree rotations around each axis and Gaussian noise. For this experiment, we apply our random RIP transformation with noise parameters $\delta$ (see Section \ref{rip}) from 0.75 to 0.9 in increments of 0.05, and our smooth perturbation with standard deviation 0.05 to 0.02 in increments of 0.05 (see Section \ref{interp}). We pre-train our models with the RIP transformation and perturbation and interpolation transformations, respectively. 

\paragraph{Results} Results for all experiments can be found in Figure \ref{fig:robustness}. For the first experiment, we find that our method's accuracy actually increases slightly with the rotation angle, unlike Figure 7 of \citet{shi2020unsupervised}, where the accuracy degrades as the rotation angle increases. We also find that our method achieves higher accuracy on the robustness experiment than the best unsupervised baseline \citet{shi2020unsupervised} at all rotation angles. In the Gaussian noise experiment we find that our method experiences only a slight decrease of around 2\% from the setting without noise to the highest level of noise, unlike Figure 8 of \citet{shi2020unsupervised}, where the accuracy decreases significantly as the noise level increases. \citet{shi2020unsupervised} achieves robustness by learning their representations by mapping the distribution of points to the corresponding point origin, but our method achieves much better robustness through a much stronger constraint of isometry-invariance on the representations. For our proposed transformations, we find similar results as the noise experiment, with only slight decreases in performance as the noise increases, showing that our method is even robust to much more complex variations. The lower accuracy of the robust RIP transformation compared to the non-robust accuracy (see Table \ref{table:single-aug}) is to be expected because \citet{zhao2020isometry} observes that robustness to random rotations causes a significant decrease in classification accuracy for supervised training, and the RIP transformation is a generalization of random rotations. 



\section{Conclusion}

In this paper we introduce a contrastive learning framework to learn isometry and almost-isometry invariant shape representations, together with novel isometric and almost-isometric data augmentations. We show empirically that our contrastive learning and isometry approach improves over previous methods in both representation effectiveness and robustness, as well as that our novel data augmentations produce much better representations using contrastive learning than existing point cloud data augmentations.   



\begin{acknowledgements} 
    The authors would like to acknowledge Joy Hsu, Jen Weng, Julia Gong for helpful discussions, Joy Hsu for suggesting the title of the paper, and Yi Shi for help with his code for baseline experiments.  

\end{acknowledgements}

\bibliography{references}

\newpage

\appendix

\section{Euclidean isometries are orthogonal matrices}
\label{orth-info}

The isometries of $n$-dimensional Euclidean space are described by the Euclidean group $E(n)$, the elements of which are arbitrary combinations of rotations, reflections, and translations. One way to describe this structure mathematically is that the group $E(n) = O(n) \rtimes T(n)$ is the semi-direct product of the group of $n$-dimensional orthogonal matrices $O(n)$ by the group of $n$-dimensional translations $T(n)$. For the purpose of learning representations from point clouds, it suffices to only consider the non-translation components of $E(n)$ since we can always normalize input point clouds, which has the effect of centering all point clouds at the origin. Mathematically, this is achieved by taking the quotient of $E(n)$ by the translation group $T(n)$, so it suffices to work only with the orthogonal group $O(n) \cong E(n)/T(n)$.

\section{RIP matrices}
\label{rip-info}

Here we provide additional characterizations of RIP matrices in terms of the spectral norm and 2-norm. We will find it easier to work with the following definition of RIP matrices: 

\begin{definition}[Adapted from \citet{zhao2020isometry}]
For all $s$-sparse vectors $x \in \mathbb{R}^n$, that is vectors $x$ with at most $s$ non-zero coordinates, matrix $A$ satisfies $s$-restricted isometry with constant $\delta$ if 
\begin{align}
    (1 - \delta)\norm{x}^2 \le \norm{Ax}^2 \le (1 + \delta)\norm{x}^2
\end{align}
\end{definition}
To see why it makes sense to describe matrices satisfying the RIP condition as  almost-orthogonal, we will follow the argument of \citet{zhao2020isometry}. In our case, our vectors will not be sparse, so we will have $s$ equal to the size of the vector $n$.
Then we can rewrite this condition as
\begin{align} \label{eqn:rip-2}
    \left| \frac{\norm{Ax}^2}{\norm{x}^2} - 1 \right| \le \delta, \forall x \in \mathbb{R}^n
\end{align}
Since $\norm{A}_2 = \sigma(A) $, where $\sigma(A)$ is the spectral norm of $A$; that is, the largest singular value of $A$. Using the min-max characterization of singular values, we know that
\begin{align}
    \sigma(A^TA - I) = \max_{x \neq 0} \frac{x^T(A^TA - I)x}{\norm{x}^2}
\end{align}
and simplifying we get
\begin{align}
    \sigma(A^TA - I) = \max_{x \neq 0} \frac{\norm{Ax}^2}{\norm{x}^2} - 1
\end{align}
Plugging this in to Equation \ref{eqn:rip-2}, we get
\begin{align}\label{eqn:rip-spectral-cond}
    \sigma(A^TA - I) \le \delta 
\end{align}
\noindent
From this equation, we can see that RIP matrices are almost-orthogonal, and therefore almost-isometric, with respect to the spectral norm. 

\section{Hyperparameter sensitivity}



\begin{figure*}
\centering
\begin{minipage}{.42\linewidth}
  \centering
  \subfloat[]{\includegraphics[width=0.5\linewidth]{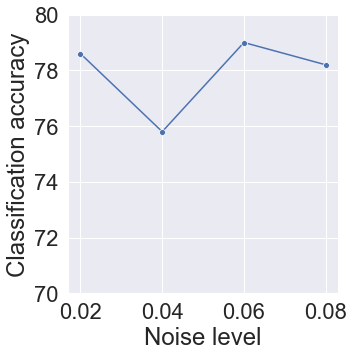}}
  \subfloat[]{\includegraphics[width=0.5\linewidth]{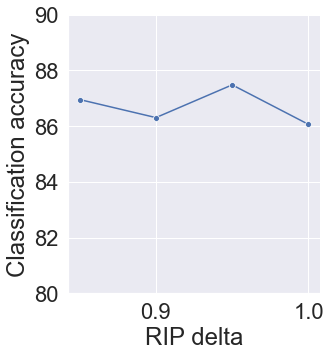}}
  \captionof{figure}{Hyperparameter sensitivity plots for (a) $\sigma$, the standard deviation of Gaussian noise in the random Gaussian perturbation augmentation, and (b) $\delta$, the deviation from isometry for our random RIP augmentation. We find that our model is not particularly sensitive to either hyperparameter.}
  \label{fig:sensitivity}
\end{minipage}%
\hspace{.02\linewidth}
\begin{minipage}{.54\linewidth}
  \centering
  \subfloat[]{\includegraphics[width=0.4\linewidth]{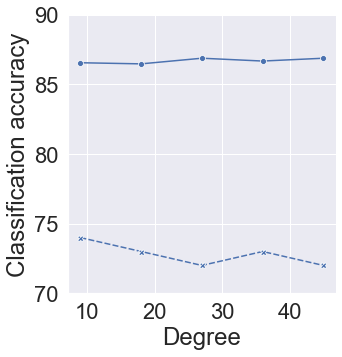}}
  \subfloat[]{\includegraphics[width=0.6\linewidth]{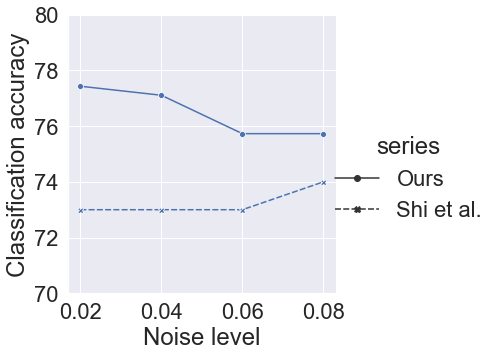}}
  \captionof{figure}{Plots of accuracy vs variation strength for (a) rotations by a fixed angle, (b) Gaussian noise of varying standard deviations for the baseline \cite{shi2020unsupervised}. We see that the method is fairly robust but less accurate than our method. One caveat is that we were unable to fully reproduce their results using their publicly available code.}
  \label{fig:shi}
\end{minipage}
\end{figure*}

We investigate the sensitivity of our model to the Gaussian noise parameter (standard deviation) $\sigma$ for Gaussian perturbations and the stretching parameter $\delta$ for RIP matrices. Results can be found in Figure \ref{fig:sensitivity}. We find that the performance of our model is not heavily effected by the choice of either parameter.

\section{Robustness comparison to baseline}

Results for the rotation and Gaussian perturbation robustness experiments on ModelNet40 of Section~\ref{robustness} using the baseline method \citep{shi2020unsupervised} can be found in Figure~\ref{fig:shi}. An identical experiment was carried out in their paper, except the classification part (see Section~\ref{shapeclassificationprotocol}) was carried out on ShapeNet instead of ModelNet40. The experiments were carried out using their publicly available implementation here: \url{https://github.com/WordBearerYI/Unsupervised-Deep-Shape-Descriptor-with-Point-Distribution-Learning}. We find that differing amounts of Gaussian noise do not affect the classification accuracy, contrary to their results on ShapeNet where as increasing rotations have a slight negative effect on classification accuracy, which reflects their ShapeNet results. We note that we were unable to reproduce their result in Table~\ref{table:prior-work} with their code. With the results we were able to produce, we find that our model has similar robustness but much better accuracy than \cite{shi2020unsupervised}. We will also make our code publicly available. 

\section{PointNet encoder architecture}

\begin{table*}[htb!]
\caption{The PointNet encoder architecture used for all versions of our model. Each layers is followed by a batch normalization layer and a ReLU layer except for the last two linear layers. The identity is added to the third linear layer as in \cite{qi2017pointnet}, and the output is reshaped at the before the second block of 1D convolutions. $C$ is the number of classes for classification.}
\vskip 0.15in
\begin{center}
\begin{small}
\begin{sc}
\begin{tabular}{c c c c c}
\hline
Layer Type & In channels & Kernel size & Stride & Out channels \\
\hline
Conv1D & 3 & 1 & 1 & 64 \\
Conv1D & 64 & 1 & 1 & 128 \\
Conv1D & 128 & 1 & 1 & 1024 \\
Linear & 1024 & -- & -- & 512 \\
Linear & 512 & -- & -- & 256 \\
Linear & 256 & -- & -- & 9 \\
\hline
Conv1D & 3 & 1 & 1 & 64 \\
Conv1D & 64 & 1 & 1 & 128 \\
Conv1D & 128 & 1 & 1 & 1024 \\
Linear & 1024 & -- & -- & $C$ \\
\hline
\end{tabular}
\end{sc}
\end{small}
\end{center}
\vskip -0.1in
\label{table:pointnet-encoder}
\end{table*}

A exact specification of our PointNet \citep{qi2017pointnet} encoder architecture can be found in Table~\ref{table:pointnet-encoder}. 

\section{Examples of transformations}

In Figure \ref{fig:examples-appendix} we provide additional examples of randomly sampled transformations from each of our proposed data augmentation methods, which are the uniform orthogonal transformation, random RIP transformation, and smooth perturbation transformation.

\section{Failure cases}

\begin{figure*}[htb!]
\vskip 0.2in
\begin{center}
\begin{minipage}{\linewidth}
    \subfloat[]{\includegraphics[width=0.25\linewidth] {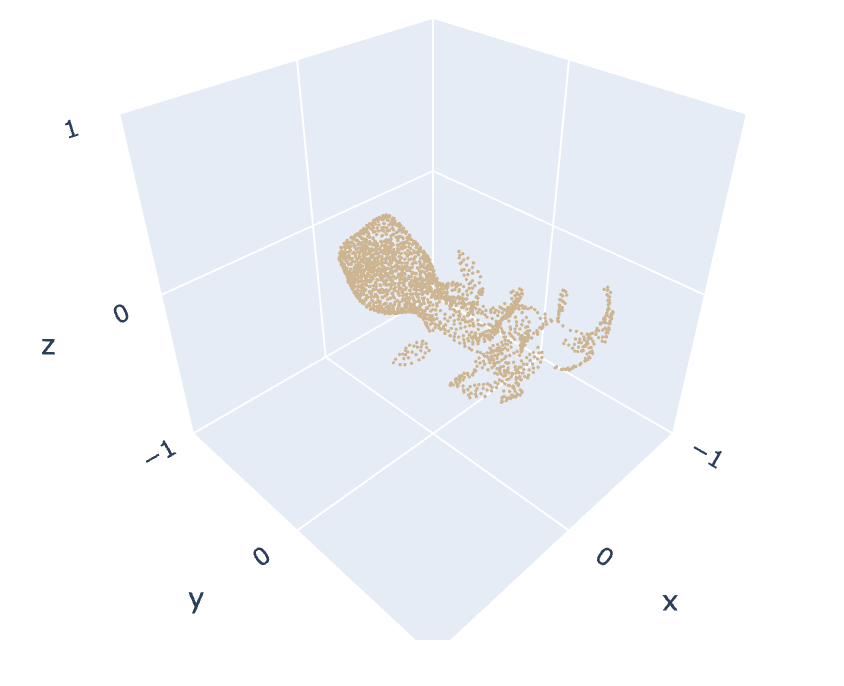}}
    \subfloat[]{\includegraphics[width=0.25\linewidth] {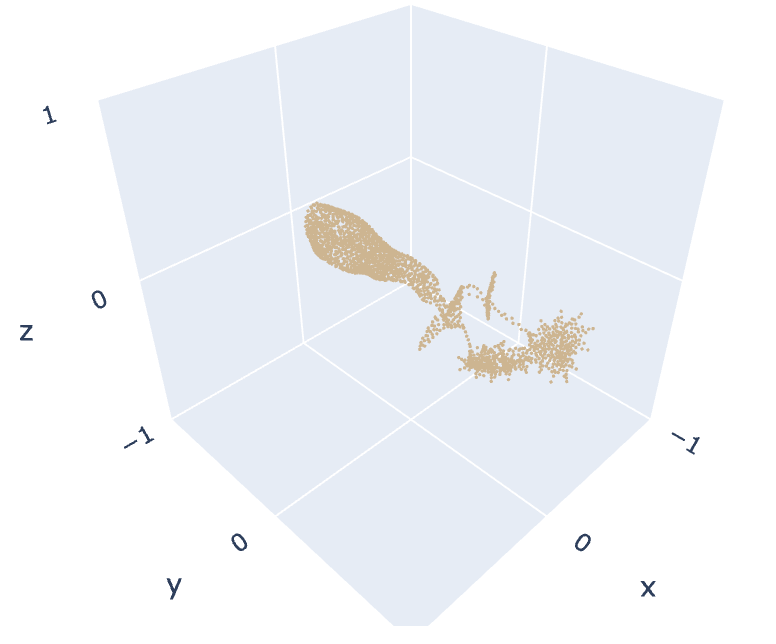}}
    \subfloat[]{\includegraphics[width=0.25 \linewidth] {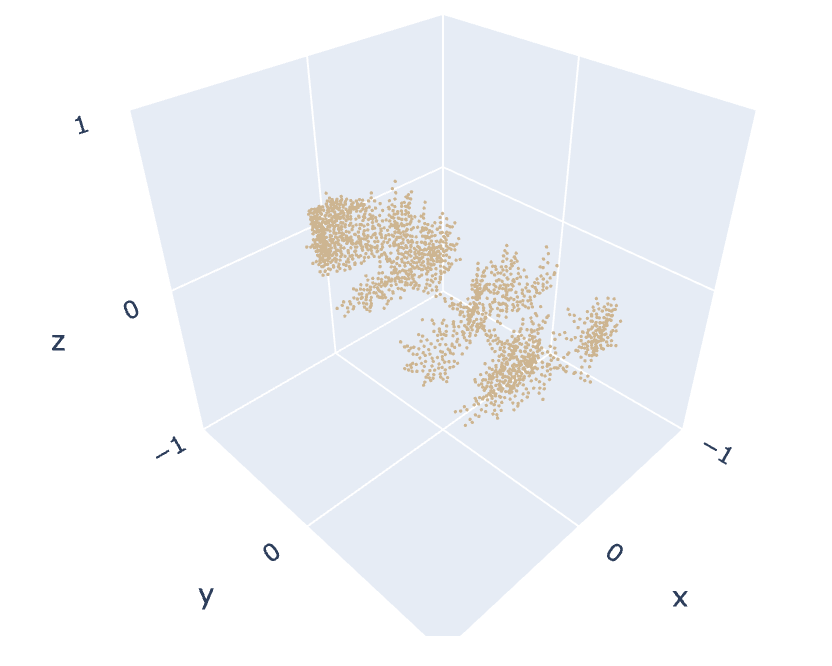}} 
    \subfloat[]{\includegraphics[width=0.25\linewidth] {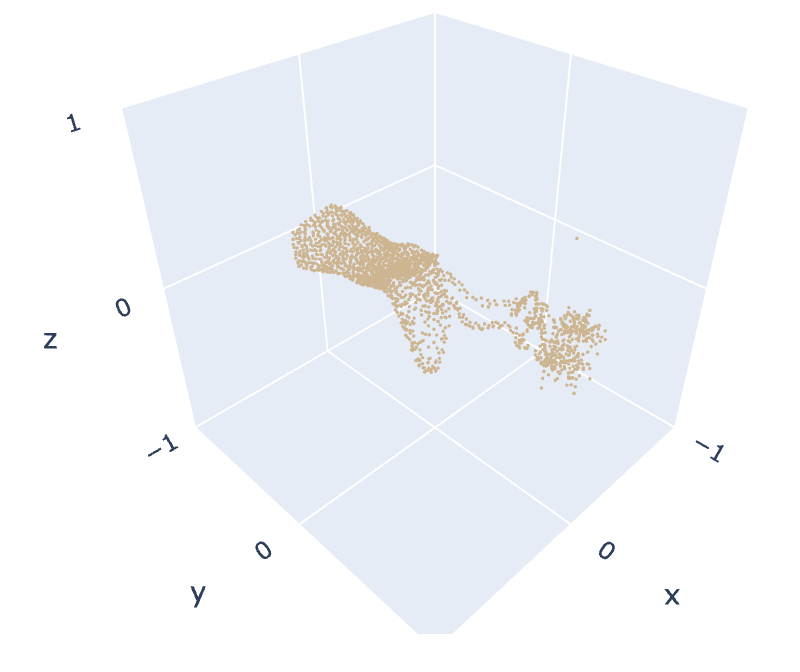}} 
\end{minipage}
\caption{(a) and (b) are examples of the flower pot class that are misclassified by our method as the plant class, and (c) and (d) are similar looking examples from the plant class.}
\label{fig:failure-cases}
\end{center}
\vskip -0.2in
\end{figure*}

In Figure~\ref{fig:failure-cases} we show examples from ModelNet40 that were misclassified by our method, and similar examples from the class it was misclassified as. The highest error rate ModelNet40 class is the flower pot class, which has an error rate much higher than any other class. Our method frequently mistakes the examples from the flower pot class for the plant class, which is much larger, and more rarely as other classes. As shown in Figure~\ref{fig:failure-cases}, examples from one class can be very similar visually to an example from another class, and we believe that this similarity is challenging for contrastive learning algorithms.

\begin{figure*}[htb!]
\vskip 0.2in
\begin{center}
\includegraphics[width=0.23\linewidth] {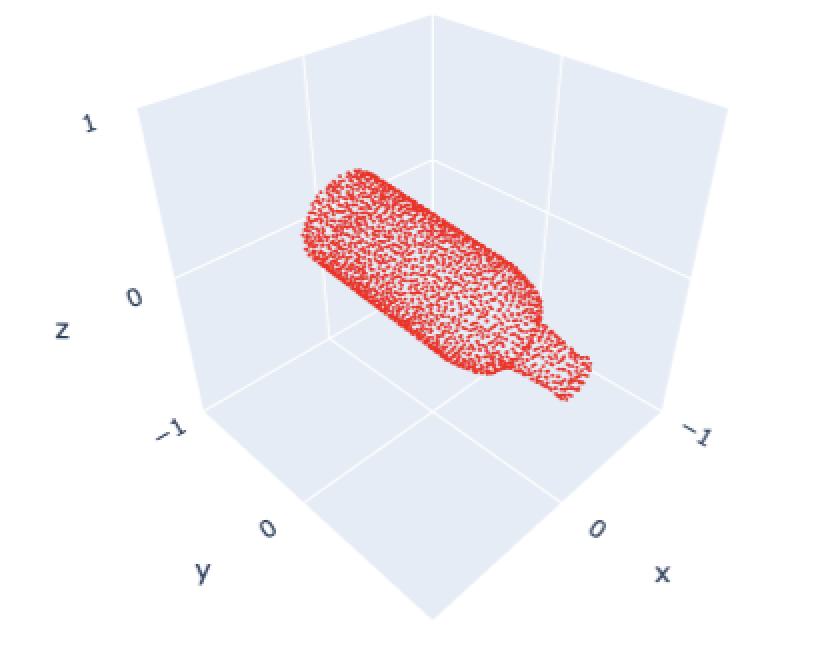}
\includegraphics[width=0.23\linewidth] {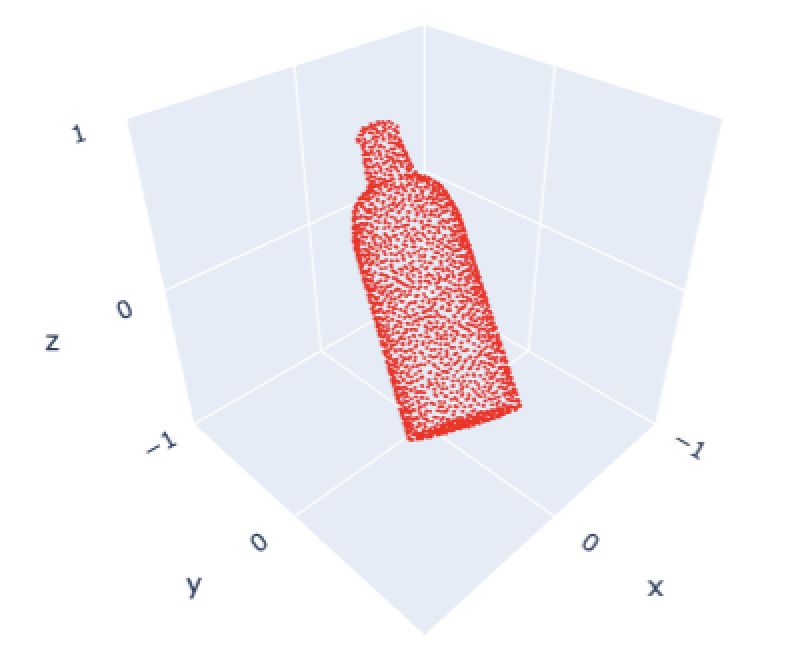}
\includegraphics[width=0.23 \linewidth] {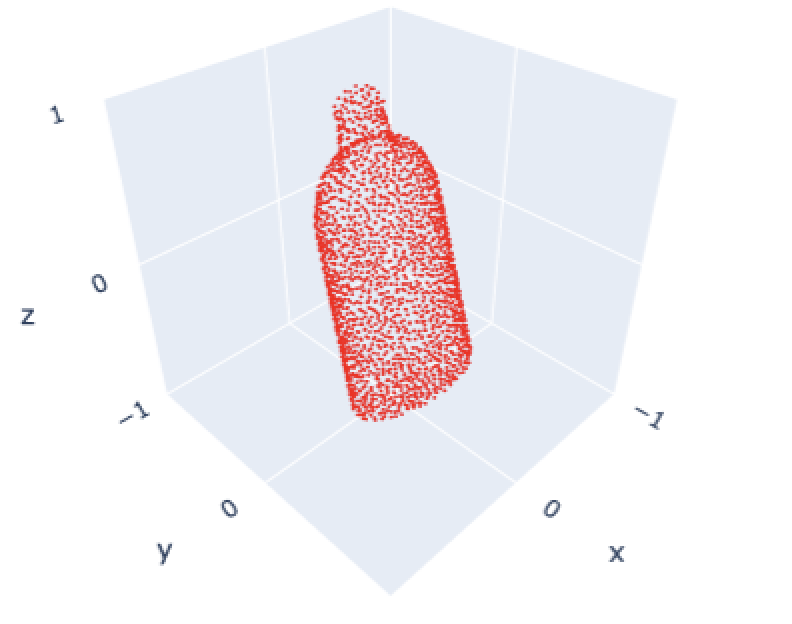}
\includegraphics[width=0.23 \linewidth] {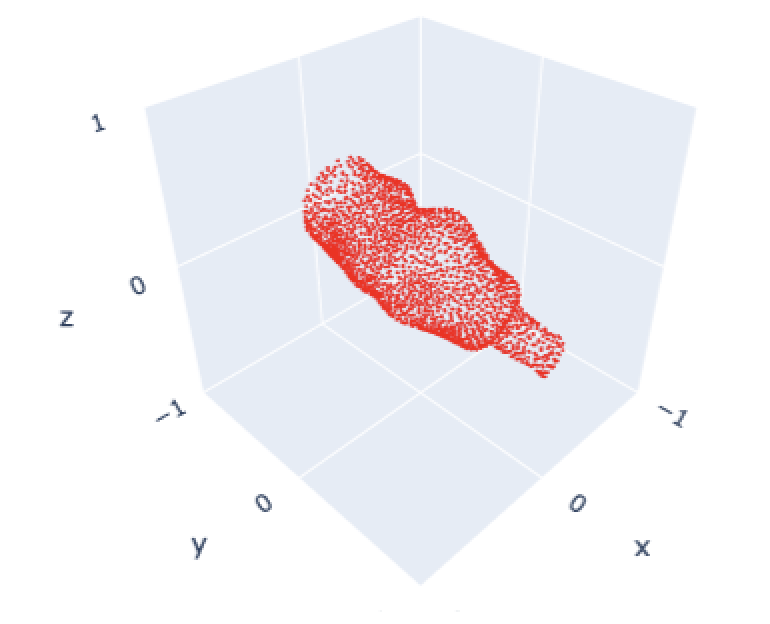}
\includegraphics[width=0.23\linewidth] {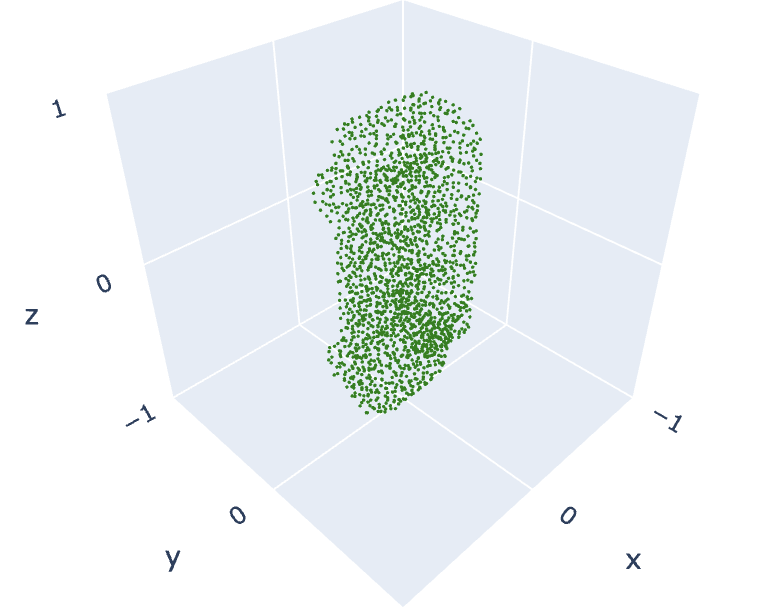}
\includegraphics[width=0.23\linewidth] {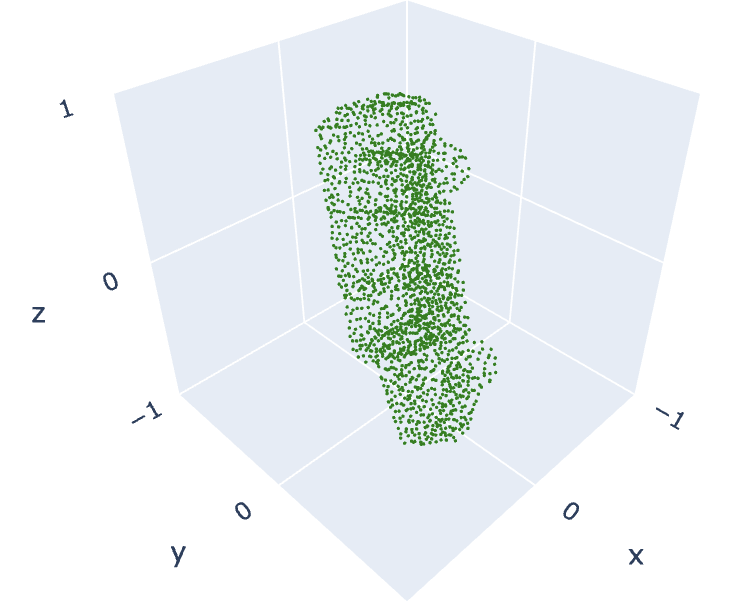}
\includegraphics[width=0.23 \linewidth] {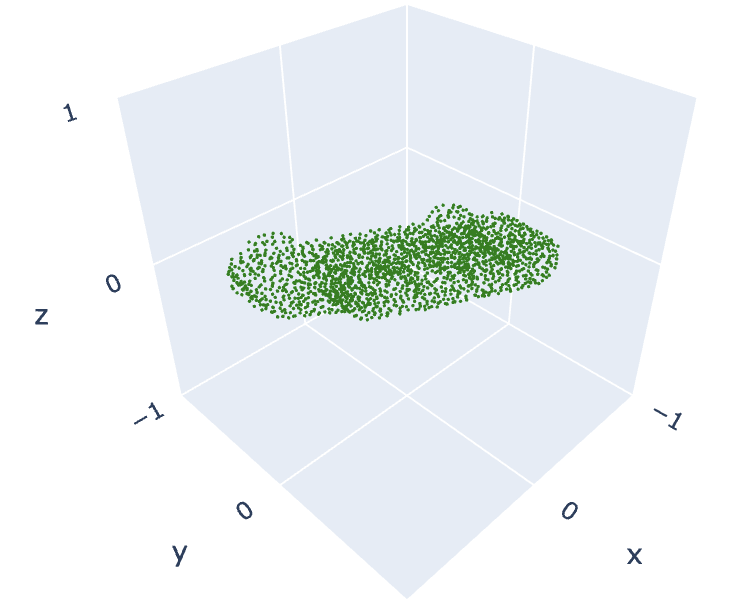}
\includegraphics[width=0.23 \linewidth] {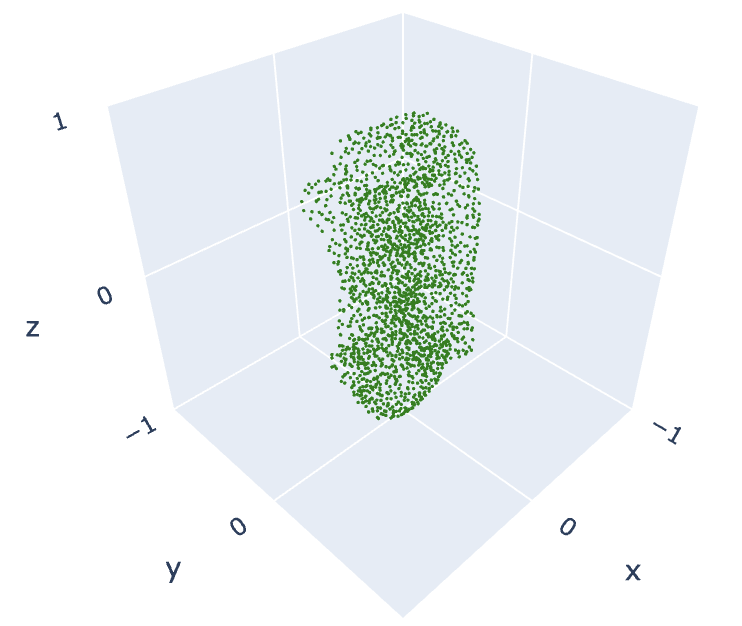}
\includegraphics[width=0.23\linewidth] {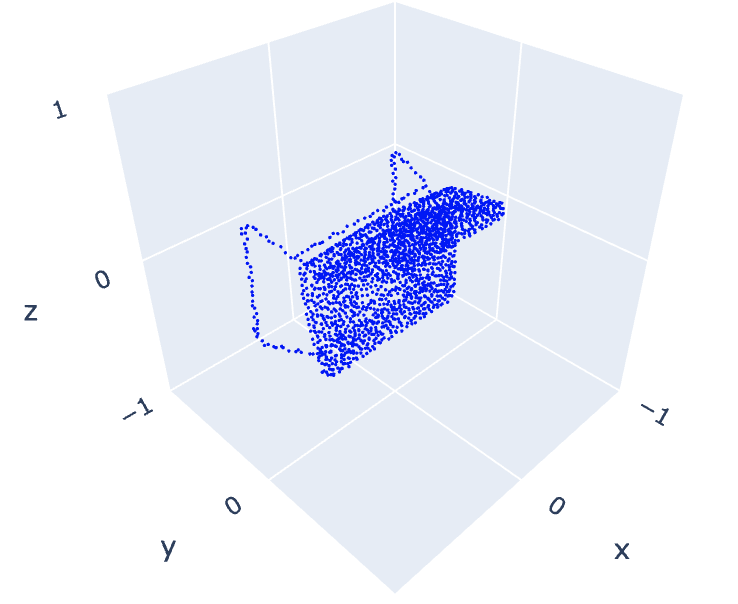}
\includegraphics[width=0.23\linewidth] {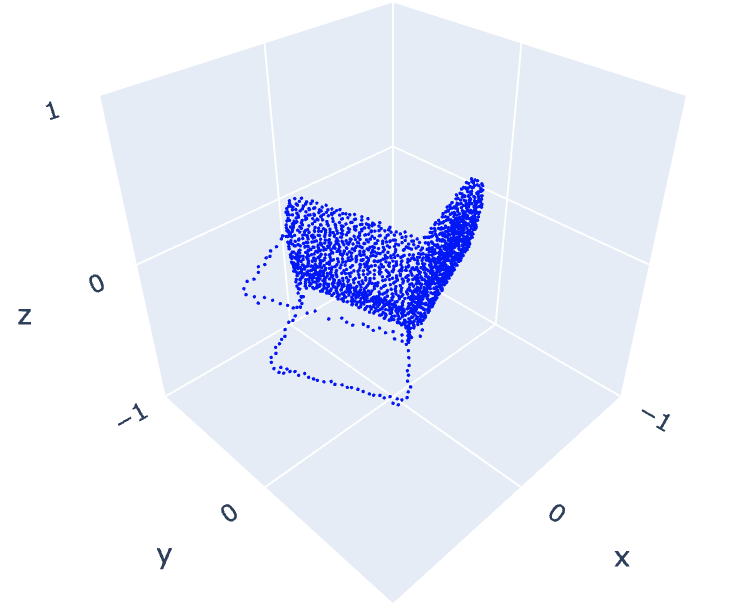}
\includegraphics[width=0.23 \linewidth] {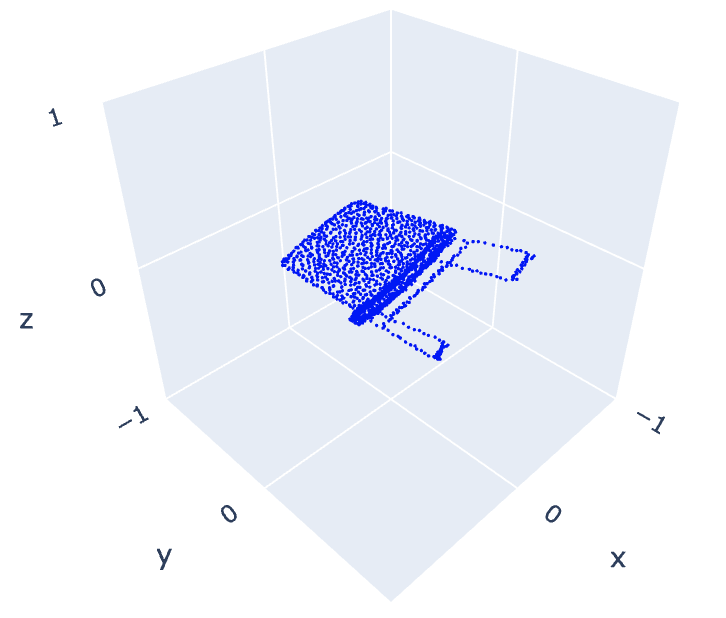}
\includegraphics[width=0.23 \linewidth] {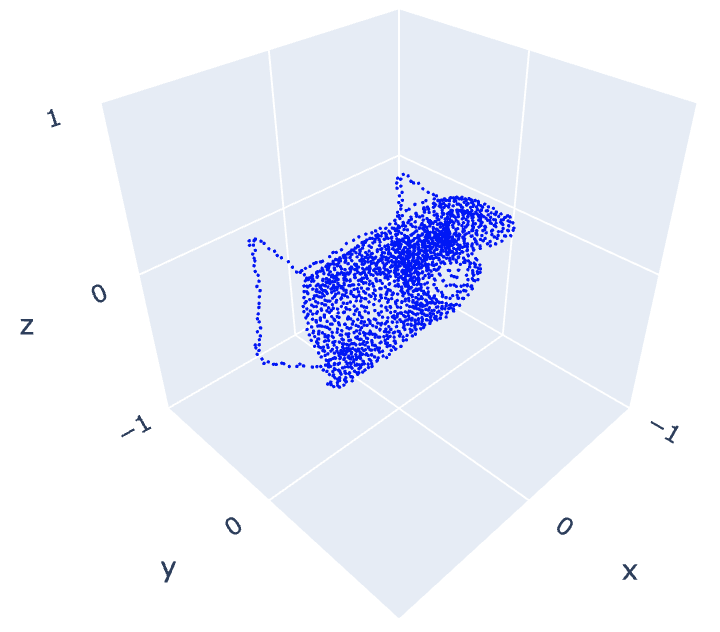}
\includegraphics[width=0.23\linewidth] {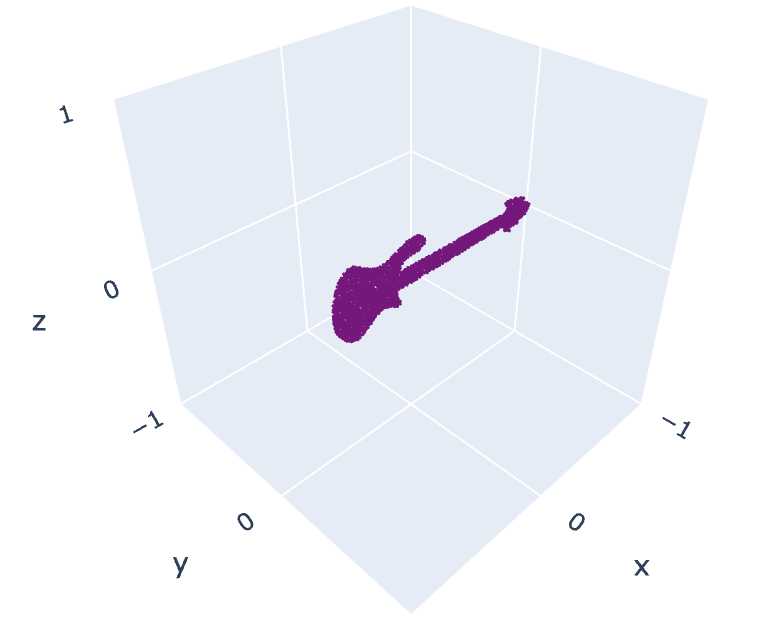}
\includegraphics[width=0.23\linewidth] {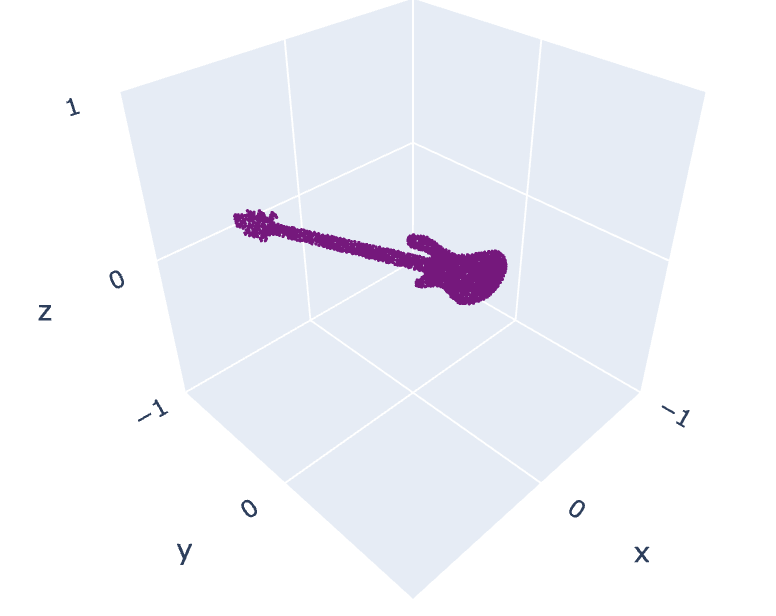}
\includegraphics[width=0.23 \linewidth] {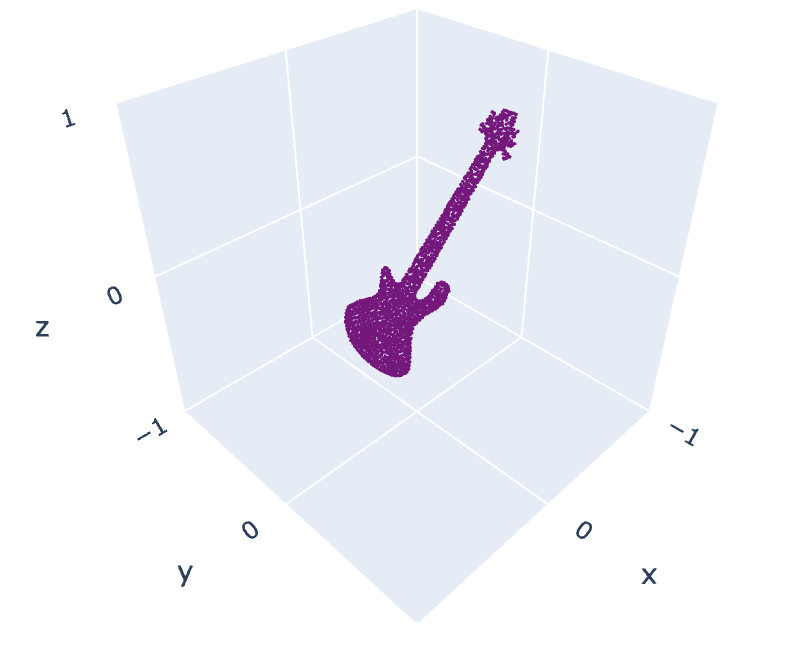}
\includegraphics[width=0.23 \linewidth] {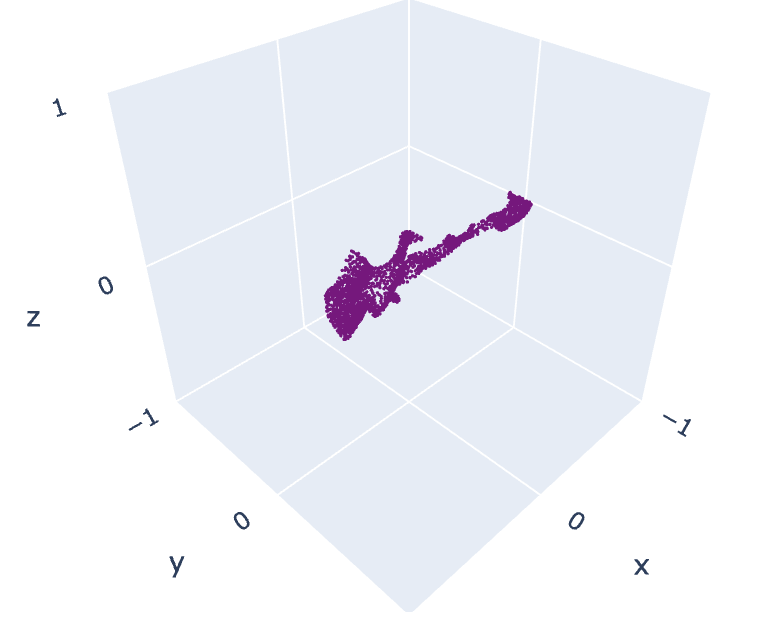}
\includegraphics[width=0.23\linewidth] {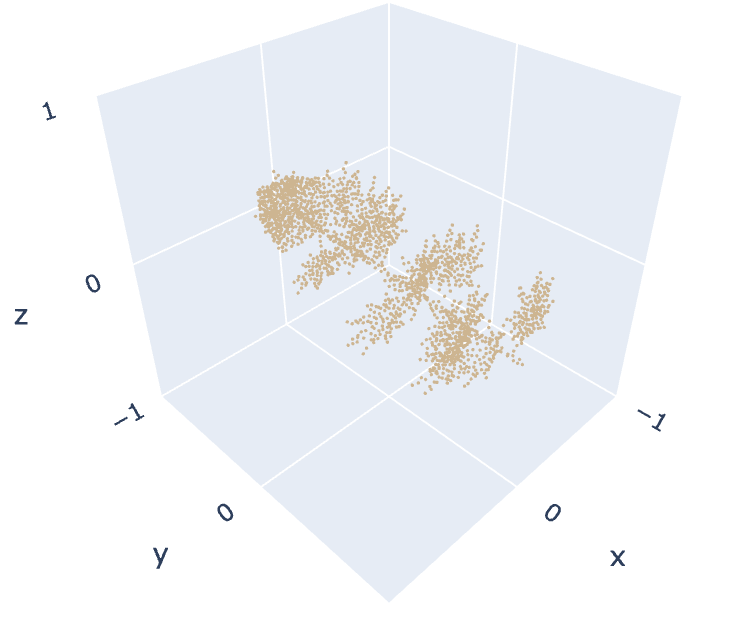}
\includegraphics[width=0.23\linewidth] {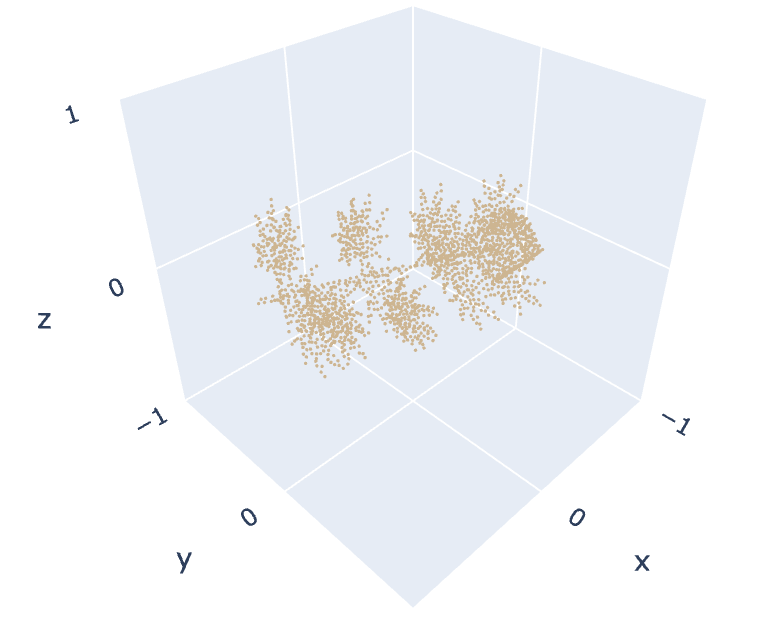}
\includegraphics[width=0.23 \linewidth] {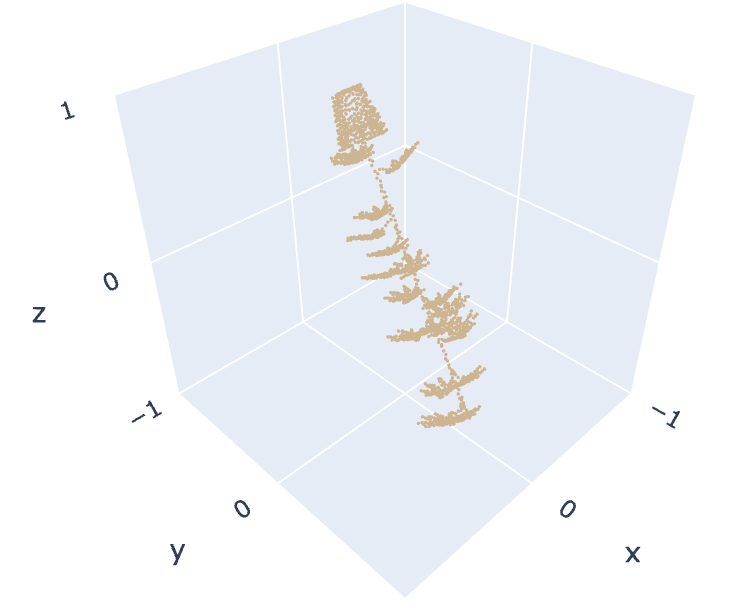}
\includegraphics[width=0.23 \linewidth] {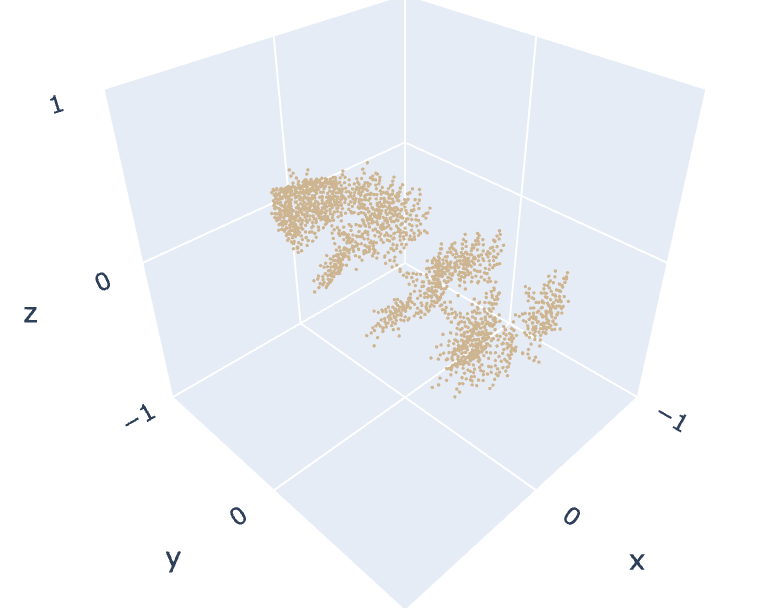}
\caption{Additional examples of randomly sampled uniform orthogonal, random RIP, and smooth perturbation transformation using our methods. In the first column from the left is the original image. In the second, third, and fourth columns from the right, we apply a randomly sampled orthogonal, RIP, and smooth perturbation transformation, respectively. We see that in general that the orthogonal transform rotates and possibly reflects the object, that the RIP transform generally rotations and slightly elongates the object, and that the smooth noise smoothly deforms the objects.}
\label{fig:examples-appendix}
\end{center}
\vskip -0.2in
\end{figure*}

\end{document}